\documentclass[conference]{IEEEtran}
\usepackage[T1]{fontenc}
\usepackage{aecompl}
\IEEEoverridecommandlockouts
\usepackage{cite}
\usepackage{amsmath,amssymb,amsfonts, bm, nccmath}
\usepackage[linesnumbered, ruled, vlined]{algorithm2e}

\SetCommentSty{mycommfont}
\usepackage{graphicx}
\usepackage{subfigure}
\usepackage{textcomp}
\usepackage{xcolor}
\usepackage{csquotes}
\let\oldnl\nl
\newcommand{\nonl}{\renewcommand{\nl}{\let\nl\oldnl}}
\DeclareMathAlphabet\mathbfcal{OMS}{cmsy}{b}{n}

\usepackage{multirow}

\begin{document}
	
	\title{Clustering-based Multitasking Deep Neural Network for Solar Photovoltaics Power Generation Prediction
		\thanks{$^*$Corresponding author: miaozheng@hdu.edu.cn}
	}
	
	\author{\IEEEauthorblockN{Hui Song, Zheng Miao$^*$}
		\IEEEauthorblockA{\textit{School of Cyberspace} \\
			\textit{Hangzhou Dianzi University}\\
			Hangzhou, China \\
			hui.song@hdu.edu.cn\\
			miaozheng@hdu.edu.cn}
		\and
		\IEEEauthorblockN{Ali Babalhavaeji, Saman Mehrnia, Mahdi Jalili, Xinghuo Yu}
		\IEEEauthorblockA{\textit{School of Engineering} \\
			\textit{RMIT University}\\
			Melbourne, VIC, Australia\\
			ali.babalhavaeji@student.rmit.edu.au, Mehrnia.saman1992@gmail.com \\
			mahdi.jalili@rmit.edu.au, xinghuo.yu@rmit.edu.au
		}
	}
	
	\maketitle
	
	\begin{abstract}
		
		The increasing installation of Photovoltaics (PV) cells leads to more generation of renewable energy sources (RES), but results in increased uncertainties of energy scheduling. Predicting PV power generation is important for energy management and dispatch optimization in smart grid. However, the PV power generation data is often collected across different types of customers (e.g., residential, agricultural, industrial, and commercial) while the customer information is always de-identified. This often results in a forecasting model trained with all PV power generation data, allowing the predictor to learn various patterns through intra-model self-learning, instead of constructing a separate predictor for each customer type. In this paper, we propose a clustering-based multitasking deep neural network (CM-DNN) framework for PV power generation prediction. K-means is applied to cluster the data into different customer types. For each type, a deep neural network (DNN) is employed and trained until the accuracy cannot be improved. Subsequently, for a specified customer type (i.e., the target task), inter-model knowledge transfer is conducted to enhance its training accuracy. During this process, source task selection is designed to choose the optimal subset of tasks (excluding the target customer), and each selected source task uses a coefficient to determine the amount of DNN model knowledge (weights and biases) transferred to the aimed prediction task. The proposed CM-DNN is tested on a real-world PV power generation dataset and its superiority is demonstrated by comparing the prediction performance on training the dataset with a single model without clustering.
	\end{abstract}
	
	
	\section{Introduction}\label{introduction}
	Solar photovoltaics (PV) energy has gained significant attraction as an alternative energy development solution, given that it is sustainable and environmentally friendly, and offers economic benefits~\cite{markovics2022comparison}. Higher PV power penetration into the smart grid and power system leads to reduced conventional energy demand and energy costs for customers~\cite{song2023multi}. Due to internal factors such as PV cell size, types of solar cells, and brands of solar panels\footnote{{https://www.energysage.com/solar/average-solar-panel-size-weight/}}, the power generation varies significantly. Weather and geographic location, as the external factors, also have impacts on PV power generation~\cite{li2021multi}. Accurate forecasting of solar PV power generation is essential for improving power management and dispatching optimization in the smart grid~\cite{du2023theory, alghamdi2023prediction}. 
	
	Many existing works have addressed PV power generation prediction problems with a variety of models.~Zhou~\textit{et al.}~\cite{zhou2020prediction} developed a hybrid model integrating a genetic algorithm for optimizing connection weights and biases within extreme learning machine (ELM), aimed at forecasting concurrent days. A neural network-based approach was employed to forecast PV power generation, incorporating meteorological variables such as irradiance, humidity, temperature, and wind speed in~\cite{zhang2021prediction}.~Li~\textit{et al.} constructed models using deep ELM under different weather information such as cloudy, sunny, and rainy days~\cite{li2021multi}. Two different machine learning methods were applied to investigate the correlation among various input variables, including solar PV panel temperature, ambient temperature, and relative humidity~\cite{zazoum2022solar}.~Polasek~\textit{et al.} targeted PV power prediction by using localized meteorological data and considering the error-prone nature of weather forecasts ~\cite{polasek2023predicting}. These models mainly focus on considering input features such as temperature to facilitate the forecasting accuracy. 
	
	There are some existing works that consider knowledge transfer within the predictor, with the development of multi-task optimization~\cite{qiao2022evolutionary, qiao2022dynamic}. For example, Wang~\textit{et al.} proposed an interpretable multi-prediction framework for forecasting day-ahead load and PV power~\cite{wang2024multi}. Their approach involved devising a non-parametric functional principal component analysis to discern significant patterns within daily data. This methodology aimed to augment overall performance by concurrently training load and PV power data.~Ju~\textit{et al.} formulated a multi-task learning (MTL) model to prevent the autoencoder from simply replicating the input to the output, particularly in the context of ultra-short-term PV power prediction~\cite{ju2020ultra}.~Shireen~\textit{et al.} developed an efficient model to iterative MTL that could improve prediction accuracy through sharing the common knowledge among data from several similar PV solar panels~\cite{shireen2018iterative}. A regionally distributed PV power prediction method using MTL was proposed, where long short-term memory (LSTM) was utilized to train the data over 10 different stations simultaneously~\cite{wang2021regional}. Those works either consider training PV power data with load power or PV power generation from different stations/locations via MTL to boost the forecasting performance. However, they use a single model, instead of modeling them separately and considering knowledge transfer among different predictors. Also, they ignore the influence of different scales of PV cell sizes. The probability matching was used to control the selection of each source task in~\cite{song2022multi} for energy demand prediction.~Song~\textit{et al.}~\cite{song2023multitasking} developed a multitasking recurrent neural network (RNN) model, where a coefficient was set for determining the amount of knowledge to be transferred among different prediction models. However, the coefficient value or probability is challenging to be determined, particularly when applied in different applications. 
	
	We propose a clustering-based multitasking deep neural network (CM-DNN) framework to address PV power generation forecasting problem. The entire dataset is first clustered to different groups according to the distributions. For each group of customers, a deep neural network (DNN) model is employed and trained until the prediction accuracy cannot be improved. Then based on the trained models, inter-model knowledge transfer is performed among different groups of PV customers. For a targeted forecasting task, source task selection is designed to select the subset of source tasks (excluding the aimed prediction task) to help train its DNN model. A coefficient value is used to decide the amount of the reused model knowledge (weights and biases) of each selected task. Particle swarm optimization (PSO) is employed to obtained the optimal subset of source tasks and coefficients. The contributions can be summarized as follows:
	
	\begin{itemize}
		
		\item We propose a CM-DNN framework to address the PV power prediction problem. Clustering is employed to group the PV power data into different customer types, which are then trained simultaneously through inter-model knowledge transfer.
		
		\item We compare CM-DNN with several popular time series forecasting models, each of which trains the entire dataset in a single model, to demonstrate its superiority.
		
	\end{itemize}
	
	The subsequent sections of this manuscript are structured as follows. The problem formulation is described in~Section~\ref{problem}. Section~\ref{method} illustrates the details of CM-DNN and its implementation. Data analysis and results are reported in~Section~\ref{result}. Section~\ref{conclusion} introduces the conclusions and our future work.

	\section{Problem Definition} \label{problem}
	
	Given PV power generation data $\bm{X} \in \mathbb{R}^{N \times M}$ ($N: \#\text{days}$, $M: \#\text{timestamps}$) that includes different types of customers such as residential, agricultural, industrial, and commercial, the prediction problem is traditionally addressed by using a predictor $\psi(\cdot)$ trained on the entire data. However, the data over these customers is significantly different due to the PV-system cell sizes. $\bm{X}$ is clustered into different customer types, denoted as $\bm{X} = \{\bm{X}_{1}, \bm{X}_{2}, \dots, \bm{X}_{4}\}$, $\bm{X}_{i} = \{\bm{x}_{i, 1}, \bm{x}_{i, 2},..., \bm{x}_{i, M_i}\}, \forall i \in \{1, 2, \dots, n\}, n = 4$, where $M_i$ represents the number of instances (days) in the $i$th type of customers, so that each of them can be modeled independently. Let $\{({\bm{X}}_1, \bm{y}_1), ({\bm{X}}_2, \bm{y}_2),..., ({\bm{X}}_n, \bm{y}_n)\}$ represent the training datasets for these $n$ types of customers, each task $i, i \in \{1, 2, \dots, n\}$ employs a DNN as the predictor, denoted as $\psi_i(\cdot)$, and the prediction process can be simplified as $\psi_i(\bm{X}_i; \mathcal{P}_i) \rightarrow \tilde{\bm{y}}_i$. $\mathcal{P}_i$ represents the DNN connection weights and biases set and $\tilde{\bm{y}}_i$ are the predicted PV power generation. These tasks are firstly trained individually using a gradient descent (GD)-based optimizer until converged, so that $\bm{\mathcal{P}}^* = \{\mathcal{P}_1^*, \mathcal{P}_2^*, \dots, \mathcal{P}_n^*\}$ can be achieved. Then for a customer type $i, i \in \{1, 2, \dots, n\}$, source task selection is performed to find the optimal subset $\bm{s}_i$ of source tasks. Each of the selected source task $k$ uses a coefficient $\alpha_{i,k}$ to determine its amount of DNN model information reused in the target forecasting task $i$. The $i$th DNN parameter set can be obtained via $\mathcal{P}_i^{'} =  \sum_{k \in \bm{s}_i}\alpha_{i,k}\mathcal{P}_k^*, \alpha_{i,k}\in[-1,1]$, so that $\psi_i({\bm{X}}_{i}; \mathcal{P}_i^{'}) \rightarrow \hat{\bm{y}}_i$. To obtain the final optimal $\mathcal{P}_i^{*}$ is to find the optimal $\bm{s}_i^{*}$ and $\bm{\alpha}_i^{*}$, and finally $\hat{\bm{y}}_i$ can be achieved via $\hat{\bm{y}}_i =\psi_i({\bm{X}}_{i}, \bm{\mathcal{P}}^*; \bm{s}_i^{*}, \bm{\alpha}_i^{*})$. 
	
	\begin{figure*}[ht!]
		\centering
		\subfigure[]
		{\centering\scalebox{0.65}
			{\includegraphics{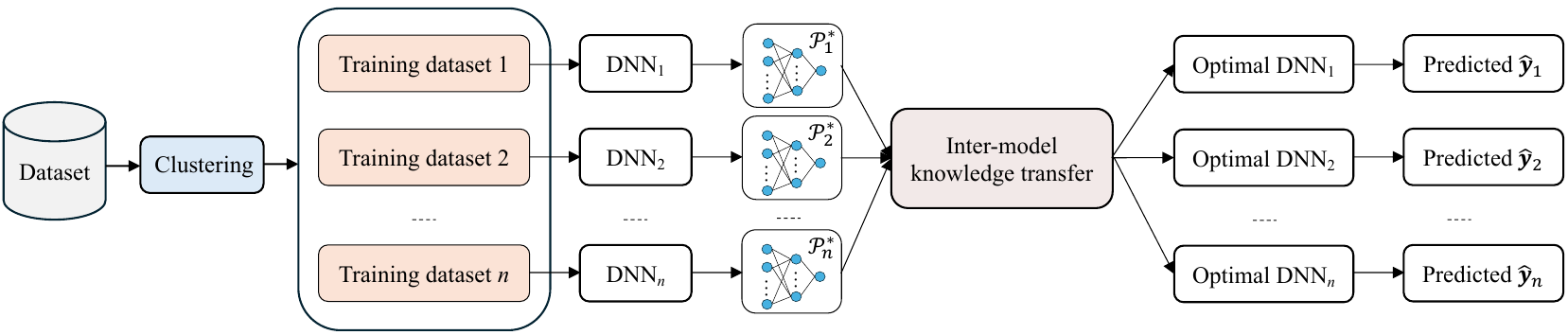}}\label{framework_1}} 
		\subfigure[]
		{\centering\scalebox{0.65}
			{\includegraphics{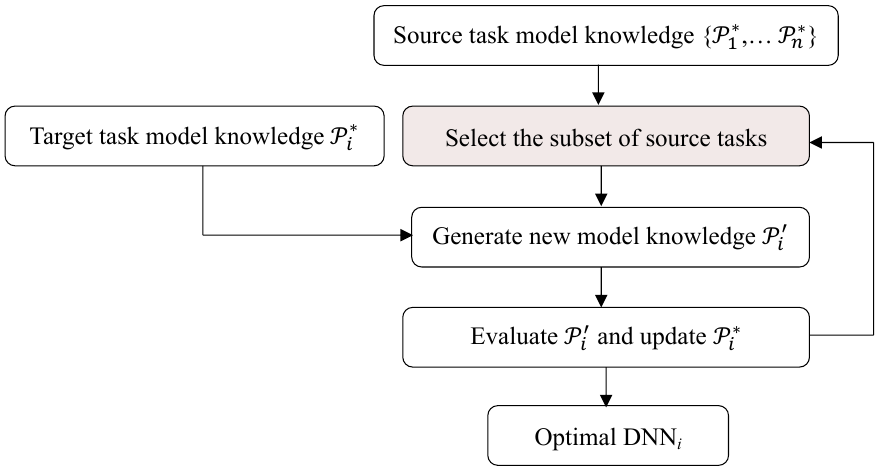}}\label{framework_2}} 
		\caption{The overall CM-DNN framework: (a) the diagram of CM-DNN working process; (b) the inter-model knowledge transfer process for a specified type of customers $i, \forall i \in \{1, 2, \dots, n\}$.}
		\label{Overall_framework}
	\end{figure*}
	
	\section{The Proposed Method}\label{method}
	
	\subsection{Framework}
	
	The proposed CM-DNN framework for PV power generation prediction is illustrated in~Fig.~\ref{Overall_framework}. The dataset $\bm{X}$ is clustered into $n$ different types of customers according to the distribution of data, as illustrated in~Fig.~\ref{framework_1}. These $n$ datasets are trained separately, each of which employs an independent DNN model. For customer type $i, i \in \{1, 2, \dots, n\}$, i.e., $\text{DNN}_i$, after obtaining the optimal parameter set $\mathcal{P}_i^*$ that includes the connection weights and biases via intra-model self-learning, inter-model knowledge transfer will be performed, as shown in~Fig.~\ref{framework_2}. During this procedure, for target task $i$, source task selection is performed to select the subset $\bm{s}_i$ from source tasks ($\mathcal{P}_i^*$ is not included), whose model knowledge will be extracted, transferred, and reused in the target task. A coefficient $\alpha_{i,k}$ is assigned to determine the amount of the $i$th DNN model knowledge in each selected task $k$ to be reused, so that a new DNN model parameter set $\mathcal{P}_i'$ for the aimed forecasting task $i$ can be generated. By comparing the performance of $\mathcal{P}_i'$ and $\mathcal{P}_i^*$, $\mathcal{P}_i^*$ will be updated. To obtain the optimal subset $\bm{s}_i^*$ of source tasks and the related coefficients $\bm{\alpha}_{i}^*$, an evolutionary algorithm (EA) is applied to address this optimization problem that consists of binary and continuous decision variables. Finally, the prediction result of the customer type $i$ can be computed through $\hat{\bm{y}}_i = \psi_i(\bm{X}_i, \bm{\mathcal{P}}^*; \bm{s}_i^*, \bm{\alpha}_{i}^*),  \bm{\mathcal{P}}^* = \{\mathcal{P}_1^*, \mathcal{P}_2^*, \dots, \mathcal{P}_n^*\}$.
	
	\subsection{Implementation}
	
	Clustering and intra-model learning will be introduced first. Then, we will present how inter-model knowledge transfer works. Finally, the details of implementing inter-model knowledge transfer are described.
	
	\subsubsection{Clustering and Intra-model Self-learning} 
	
	To category the PV power data into different groups of customers according to the distributions, K-means is applied~\cite{kanungo2002efficient}. $K$ in K-means is the same as the number of customer type $n$. Since the data is time-dependent, i.e., daytime, we use the Euclidean distance as the evaluation metrics. Given the data $\bm{X}$, $n$ instances $\{\bm{C}_1, \bm{C}_2, \dots, \bm{C}_n\}$ are randomly chosen as the centers. For each instance $\bm{x}_i$ in $\bm{X}$, the distance $d_{i,j}$ between $\bm{x}_i= \{x_{i,1}, x_{i,2}, \dots, x_{i,M}\}$ and $\bm{C}_j= \{C_{j,1}, C_{j,2}, \dots, C_{j,M}\}, j\in \{1, 2, \dots, n\}$ can be calculated via:
	\begin{align}
		\centering
		\label{K_means_distance}
		d_{i,j}	= \sqrt{\sum_{m=1}^{M}({x}_{i,m} - {C}_{j,m})^2}
	\end{align}
	
	The cluster of $\bm{x}_i$ can be obtained via $j^* = \arg \min \bm{d}_i, \bm{d}_i = \{d_{i,1}, d_{i,2}, \dots, d_{i,n}\}$. After finding the cluster of each instance, the center of the $j$th cluster can be updated using $\bm{C}_j = \frac{1}{c_j}\sum_{\bm{x} \in \bm{X}_j}\bm{x}$, i.e., the average of instances in the $j$th cluster, where $c_j$ represents the number of instances in the $j$th cluster. The distance $d_{i,j}$ can be recalculated using (\ref{K_means_distance}), and the cluster of $\bm{x}_i$ can be then updated. With a predefined terminal condition, $\bm{X}$ can be clustered into $\{\bm{X}_1, \bm{X}_1, \dots, \bm{X}_n\}$. 
	\begin{figure}[ht!]
		\centering
		\includegraphics[scale = 0.55]{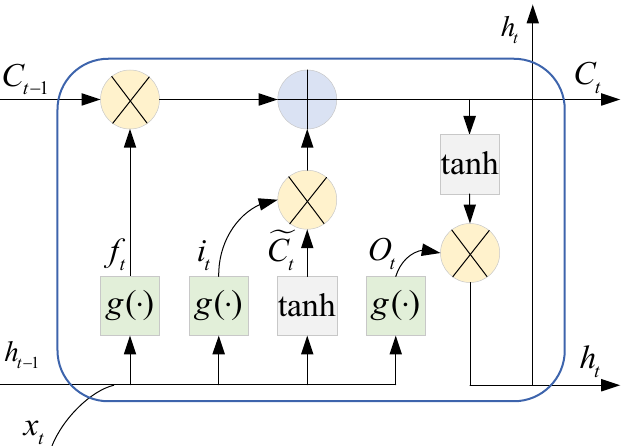}
		\caption{The structure of a cell in LSTM.}
		\label{LSTM}
	\end{figure}
	
	After clustering the entire dataset into $n$ categories, we use LSTM~\cite{hochreiter1997long} as an example to elucidate its contribution to the CM-DNN framework. The structure of a cell in LSTM is illustrated in~Fig.~\ref{LSTM}.
	
	Utilizing inputs including the current timestamp $x_t$, as well as the cell state and hidden state from the preceding timestamp $C_{t-1}$ and $h_{t-1}$, the subsequent cell and hidden states $\left(C_t,h_t\right)$ can be computed through:
	%	With inputs of the current timestamp $x_t$, the cell state and the hidden state of the previous timestamp $C_{t-1}$ and $h_{t-1}$, the next cell and hidden state $\left(C_t,h_t\right)$ can be calculated through:
	\begin{align}
		f_t & =g\left(W_f \cdot\left[h_{t-1}, x_t\right]+b_f\right) \nonumber \\
		i_t & =g\left(W_i \cdot\left[h_{t-1}, x_t\right]+b_i\right) \nonumber \\
		\tilde{C}_t & =\tanh \left(W_C \cdot\left[h_{t-1}, x_t\right]+b_C\right) \nonumber \\
		C_t & =f_t \otimes C_{t-1}+i_t \otimes \tilde{C}_t  \nonumber\\
		O_t & =g\left(W_o\left[h_{t-1}, x_t\right]+b_o\right) \nonumber\\
		h_t & =O_t \otimes \tanh \left(C_t\right)
		\label{LSTM_process}
	\end{align}
	
	With the obtained next hidden state $h_t$, the predicted values $\hat{y}_t$ is obtained through: 
	\begin{equation}
		\hat{y}_t=g\left(W_y h_t+b_y\right) \label{LSTM_prediction}
	\end{equation}
	where $g(\cdot)$ represents the Sigmoid function and  the operation $\otimes$ denotes the element-wise vector product. The parameter set $\mathcal{P} = \left\{W_f, W_i, W_C, W_o, W_y, b_f, b_i, b_C, b_o, b_y\right\}$  denotes the weight matrices and bias vectors that constitute the knowledge information in LSTM. $C_t$ serves for the vanishing gradient issue to retain long-term dependencies, while $h_t$ enables the network to navigate and make complex decisions over short periods of time. The simplification of the LSTM learning process from (\ref{LSTM_process}) to (\ref{LSTM_prediction}) can be expressed as $\hat{\bm{y}} =  \psi(\bm{X}; \mathcal{P})$. To evaluate the accuracy of $\hat{\bm{y}}$, root mean square error (RMSE) is used, so that the performance $\mathcal{L}$ of the prediction can be calculated via: 
	\begin{align}
		\centering
		\mathcal{L}(\bm{y}, \hat{\bm{y}})&=\sqrt{\frac{1}{ST}\sum_{s=1}^{S}\sum_{t=1}^{T}(y_{s,t} - \hat{y}_{s,t})^2}\label{loss_function_refined} \\
		\mathcal{P}^* &	= \operatorname*{argmin}_\mathcal{P} \mathcal{L}(\bm{y}, \hat{\bm{y}}) \label{loss_function}
	\end{align}
	where $\bm{y} \in \mathbb{R}^{S \times T}$, $S$ represents the quantity of samples, and $T$ signifies the number of steps to be forecasted within each sample. $\mathcal{L}(\bm{y}, \hat{\bm{y}})$ can be optimized with a GD-based optimizer, such as Adam~\cite{kingma2014adam}, to obtain the most accurate $\hat{\bm{y}}$. 
	
	\subsubsection{Inter-model Learning}
	
	For each trained model $i, i\in \{1, 2, \dots, n\}$ over a type of customers, after obtaining the optimal DNN architecture $\mathcal{P}_i^*$, it is possible to further enhance its accuracy through inter-model knowledge transfer. In this process, trained models catering to different types of customers can exchange optimal connection weights and biases. However, some models may not contribute effectively to the target prediction task. Different from~\cite{song2023two} that uses all source tasks, for a specified forecasting problem $i$, i.e., customer type $i$, source task selection is performed to choose task (exclude the target type of customers) subset $\bm{s}_i$ to aid in training the DNN for the target task. A coefficient $\alpha_{i,k}$ is used to control the amount of DNN model information from each selected task $k$, so that the newly generated parameter set $\mathcal{P}_i'$ of the target forecasting task $i$ can be calculated according to:
	\begin{align}
		\centering
		\mathcal{P}_i' = \sum_{k \in \bm{s}_i}\alpha_{i,k}\mathcal{P}_k^* + \alpha_{i,i}\mathcal{P}_i^*, \alpha_{i,k} \in [-1, 1]\label{coefficient}
	\end{align}
	The final optimal $\mathcal{P}_i^*$ can be obtained via optimizing the subset $\bm{s}_i$ and coefficients $\bm{\alpha}_i, \forall k \in \bm{s}_i, \alpha_{i,k} \in \bm{\alpha}_i$.
	
	\subsubsection{Particle Swarm Optimization}
	
	PSO~\cite{kennedy1995particle}, proposed by Kennedy~\textit{et al.} in 1995, is a population-based optimization algorithm that aims to address complex continuous, discrete, convex, nonconvex, and large-scale optimization problems~\cite{song2019multitasking}. For an optimization problem with $D$ decision variable, $ps$ particle' positions $\bm{U} = \{\bm{u}_1, \bm{u}_2, \dots, \bm{u}_{ps}\}$ and velocities $\bm{V} = \{\bm{v}_1, \bm{v}_2, \dots, \bm{v}_{ps}\}$ are initialized, $\forall p\in \{1, 2, \dots, ps\}, \bm{u}_p\in \mathbb{R}^{1 \times D}, \bm{v}_p\in \mathbb{R}^{1 \times D}$. $\bm{u}_p$ and $\bm{v}_p$ in the $g$th generation are updated according to:
	\begin{align}
		\centering
		\bm{v}_p^g & = \omega \bm{v}_p^{g-1} + c_1r_1(\bm{pbest}_p - \bm{u}_p^{g-1}) + c_2r_2(\bm{gbest} - \bm{u}_p^{g-1})  \nonumber \\
		& \qquad \bm{v}_p^g = \min(\bm{v}^{\textit{max}}, \max(-\bm{v}^{\textit{max}}, \bm{v}_p^g)) \label{velocity} \\
		& \qquad \qquad \qquad \bm{u}_p^g = \bm{u}_p^{g-1} + \bm{v}_p^g \nonumber \\
		& \qquad  \bm{u}_p^g = \min(\bm{u}^{\textit{max}}, \max(-\bm{u}^{\textit{max}}, \bm{u}_p^g))  	\label{position}
	\end{align}
	where $\bm{gbest} = \{\textit{gbest}_1, \textit{gbest}_2, \dots, \textit{gbest}_D\}$ describes the best position achieved so far over the entire population $\textit{ps}$, and $\bm{pbest}_p = \{\textit{pbest}_{p,1}, \textit{pbest}_{p,2}, \dots, \textit{pbest}_{p,D}\}$ denotes the best position achieved so far for the $p$th particle. $\omega$ is used to balance the global and local search ability. $c_1$ and $c_2$ represent the acceleration factors that motivate each particle towards its $\bm{pbest}$ and $\bm{gbest}$. $r_1$ and $r_2 \in [0, 1]$ are two random numbers. $\bm{v}^{\textit{max}}$ and $\bm{u}^{\textit{max}}$ are the maximal velocity and position, respectively.
	
	For target task $i, \forall i\in \{1, 2, \dots, n\}$, we employ PSO to optimize $\bm{s}_i$ and $\bm{\alpha}_i$.  To simplify this problem, $\bm{s}_i$ is an $n$ dimensional binary vector and $\bm{\alpha}_i$ is an $n$ dimensional vector with real values. In this case, the dimension $D = 2n$, and this problem involves decision variables consisting of binary and continuous values, and (\ref{coefficient}) can be refined as:
	\begin{align}
		\centering
		\mathcal{P}_i' = \sum_{k=1}^{n}s_{i,k}\alpha_{i,k}\mathcal{P}_k^*, \alpha_{i,k} \in [-1, 1], s_{i,k}\in\{0,1\}\label{coefficient_refined}
	\end{align}
	where $s_{i,k}=1$ represents the source task is selected, and $s_{i,k}=0$ means the source task is not selected. Specifically, we set ${s}_{i,i} = 1$ and $\alpha_{i,i} = 1$, considering that the task $i$ contributes to the new $\mathcal{P}_i'$ mostly. Since PSO cannot optimize binary decision variables directly, each solution $\bm{u}_p$ is 1D vector with continuous values in $[-1,1]$ through the evolution and the corresponding decision variables in $\bm{u}_p$ for $\bm{s}_i$ are decoded to $0/1$ with threshold $0$. 
	
	With the newly generated $\mathcal{P}_i'$, the predicted values $\hat{\bm{y}}_i = \psi(\bm{X}_i; \mathcal{P}_i') = \psi(\bm{X}_i, \bm{\mathcal{P}}^*; \bm{s}_i, \bm{\alpha}_i)$, the loss function in PSO can be formulated as:
	\begin{align}
		\centering
		\min \mathcal{L}(\bm{y}_i, \hat{\bm{y}}_i)&=\mathcal{L}(\bm{y}_i, \psi_i(\bm{X}_i, \bm{\mathcal{P}}^*; \bm{s}_i, \bm{\alpha}_i)), \forall \bm{s}_i, \bm{\alpha}_i \in \bm{u}_p \nonumber \\
		& = \mathcal{L}(\bm{y}_i, \psi_i(\bm{X}_i, \bm{\mathcal{P}}^*; \bm{u}_p)) \label{PSO_objective}
	\end{align}
	where $\psi_i(\cdot)$ represents the DNN calculation process from (\ref{LSTM_process}) to (\ref{LSTM_prediction}). With the predefined $\textit{ps}$ and $\textit{MGen}$, the implementation of PSO in addressing inter-model knowledge transfer is detailed in Algorithm.~\ref{PSO_pseudocode}. When testing CM-DNN, each predictor $\psi_i(\cdot)$ can utilize the obtained optimal $\mathcal{P}_i^*$ to calculated predicted values via (\ref{LSTM_process}) and (\ref{LSTM_prediction}).
	
	\begin{algorithm}[h!]
		\footnotesize
		\caption{Inter-model Knowledge Transfer}
		\SetAlgoLined
		\label{PSO_pseudocode}
		\KwIn{$(\bm{X}, \bm{y}) = \{(\bm{X}_1, \bm{y}_1), \bm{X}_2, \bm{y}_2), \dots, (\bm{X}_n, \bm{y}_n)\}$, $\bm{\mathcal{P}}^*$, $D, \textit{MGen}$, $ps$}  
		\For{$i \rightarrow 1: n$}{
			\nonl $\#\textit{g} = 1$ \\
			\nonl For the target customer type $i$, initialize positions $\bm{U}$ and velocities $\bm{V}$ with $\textit{ps}$ particles, set $\bm{gbest} = \bm{0}$, $\textit{gbestval} =\mathcal{L}_i$, $\bm{pbest}_p = \pmb{\emptyset}$, $\bm{Pbest}=\{\bm{pbest}_1, \dots, \bm{pbest}_{ps}\}$, $\bm{pbestval}=\{\textit{pbestval}_1, \dots, \textit{pbestval}_{ps}\} = \pmb{\emptyset}$ \\
			\For{$p \rightarrow 1: ps$}{
				\nonl Using~(\ref{PSO_objective}) to evaluate particle $p$\\
				\nonl $\bm{pbest}_p = \bm{u}_p^g$ \\
				\nonl $\textit{pbestval}_p = \mathcal{L}(\bm{y}_i, \phi_i(\bm{X}_i, \mathbfcal{P}^*; \bm{u}_p^g))$\\
				\If{$\textit{pbestval}_p < \textit{gbestval}$}{
					\nonl $\textit{gbestval}  = \textit{pbestval}_p$\\
					\nonl $\bm{gbest} = \bm{pbest}_p$}} 
			\While{$\#\textit{g} < \textit{MGen}$}{
				$\#\textit{g} = \#\textit{g} + 1$
				\For{$p \rightarrow 1: ps$}{
					\nonl According to~(\ref{velocity}) and~(\ref{position}), update $\bm{v}_p^g$ and  $\bm{u}_p^g$ \\
					\nonl Using~(\ref{PSO_objective}) to evaluate $\bm{u}_p^g$, i.e., $\mathcal{L}(\bm{y}_i, \phi_i(\bm{X}_i, \mathbfcal{P}^*; \bm{u}_p^g))$ \\
					\If{$\mathcal{L}(\bm{y}_i, \phi_i(\bm{X}_i, \mathbfcal{P}^*; \bm{u}_p^g)) < \textit{pbestval}_{p}$}{
						\nonl $\bm{pbest}_p = \bm{u}_p^g$ \\
						\nonl $\textit{pbestval}_{p} = \mathcal{L}(\bm{y}_i, \phi_i(\bm{X}_i, \mathbfcal{P}^*; \bm{u}_p^g))$}
					\If{$\textit{pbestval}_{p} < \textit{gbestval}$}{
						\nonl $\bm{gbest} = \bm{pbest}_p$ \\
						\nonl $\textit{gbestval} = \textit{pbestval}_{p}$}
				}
			}
			$\mathcal{P}_i^* = \sum_{k=1}^{n}s_{i,k}\alpha_{i,k}\mathcal{P}_k^*, \forall \bm{s}_{i}, \alpha_{i} \in \bm{gbest}$ \\
			$\hat{\bm{y}}_i = \psi_i(\bm{X}_i; \mathcal{P}_i^*)$
		}
		\KwOut{$\mathcal{P}_1^*,\dots, \mathcal{P}_n^*$, $\hat{\bm{y}}_1, \dots, \hat{\bm{y}}_n$}
	\end{algorithm}
	
	\section{Results}\label{result}
	
	The data information and experimental settings are described firstly. Then, the prediction performance on the combined dataset and clustered datasets with CM-DNN is presented and compared. Finally, the prediction results over some instances from residential, agricultural, industrial, and commercial testing datasets are visualized.

	\subsection{Data Description and Experimental Settings}
	
	%	The PV power generation data is collected at 30-minute intervals with smart meters across the CitiPower/Powercor network from January 2020 to December 2020 in Victoria, Australia. 
	
	The data for testing the proposed CM-DNN is gathered at 30-minute intervals utilizing smart meters deployed across the CitiPower/Powercor network spanning from January 2020 to December 2020 within Victoria, Australia. The dataset includes mixed customers such as residential, agricultural, industrial, and commercial. Considering that PV power is only generated during the daytime, the data from 7:00 am to 6:30 pm each day is used for modeling. The data between 6:00 am and 1:00 pm is regarded as the inputs to forecast the generation between 1:30 pm and 6:30 pm (the outputs). The total number of samples (days/instances) in the whole dataset is 26,870, with the training and testing datasets occupying $80\%$ and $20\%$, respectively. For the clustered groups, Table.~\ref{number_samples} introduces the number of samples for training and testing datasets.
	\begin{table}[ht!]
		\centering
		\caption{The number of samples for training and testing datasets across residential, agricultural, industrial, and commercial categories.}
		\begin{tabular}{c|cccc}
			\hline
			&  \textbf{Residential} & \textbf{Agricultural}  & \textbf{Industrial} & \textbf{Commercial} \\ \hline
			\textbf{Training}  & 14529       & 4513        & 1967        & 485         \\
			\textbf{Testing}   & 3633        & 1129        & 492         & 122         \\ \hline
		\end{tabular}
		\label{number_samples}
	\end{table}
	
	Since the data over each type of customer is not labeled, K-means is first applied to cluster the customers into groups of residential, agricultural, industrial, and commercial. To demonstrate the effectiveness of the proposed CM-DNN, the following comparison is performed:
	\begin{itemize}
		\item The data over all customers are regarded as a whole for modeling with one predictor, where knowledge is transferred within the model via the connection weights and biases. To ensure a fair comparison, the training data from all customer categories is combined for modeling, while the testing data is merged together for evaluating the model. 
		
		\item Each type of customer has a unique predictor that is trained with the related dataset, where inter-model knowledge transfer is performed among different tasks.
		
		\item The performance of CM-DNN is tested on some popular time series prediction models such as RNN, 1DCNN LSTM (CNN-LSTM)~\cite{livieris2020cnn}, LSTM, and Gated Recurrent Unit (GRU)~\cite{cho2014properties} over the above two cases. 
	\end{itemize}
	
	No matter for CM-DNN or the comparison models, we set the number of hidden neurons as 10. In inter-model knowledge transfer process, $\textit{ps} = 15$ and $\textit{MGen} = 100$ in PSO. For CM-DNN, we set the maximal iteration in GD-based learning to 3000, beginning with optimal DNN architecture determined by GD-based intra-model self-learning. To ensure a fair comparison, the maximal iteration for RNN, CNN-LSTM, LSTM, and GRU on the entire data without clustering is set to 4500. All prediction models use 10 hidden neurons and are independently run 10 times. 
	\begin{figure}[ht!]
		\centering
		\subfigure[]
		{\centering\scalebox{0.6}
			{\includegraphics{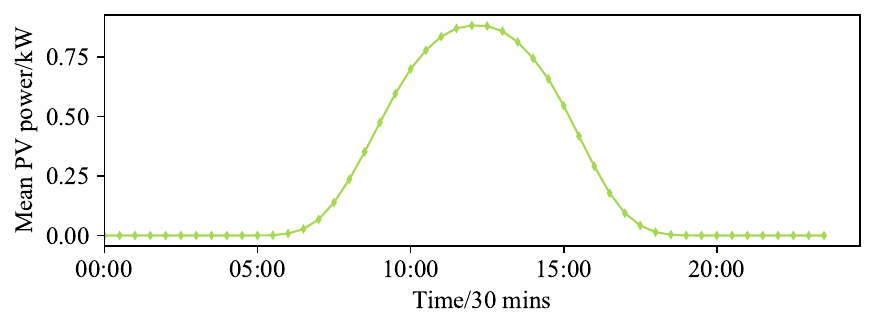}}\label{analysis_1}} 
		\subfigure[]
		{\centering\scalebox{0.6}
			{\includegraphics{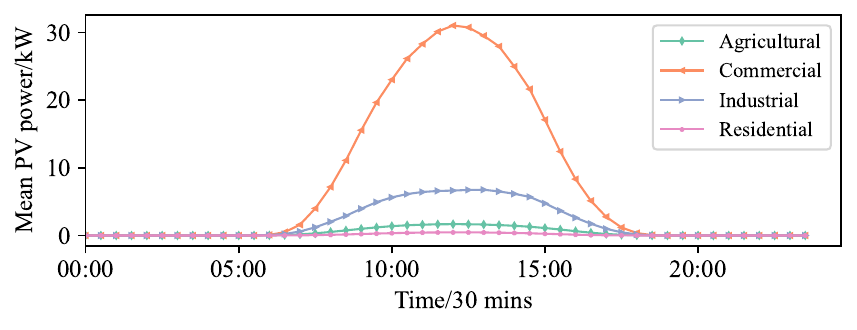}}\label{analysis_2}} 
		\caption{Clustering PV power generation into residential, agricultural, industrial, and commercial datasets.}
		\label{analysis}
	\end{figure}
	
	\subsection{Data Analysis and Clustering}
	The average PV power generation (kW) across all types of customers in the entire dataset is illustrated in~Fig.~\ref{analysis_1}, which includes all types of customers. Since the customer type is not labeled, we use K-means to cluster the entire dataset into residential, agricultural, industrial, and commercial datasets, as shown in~Fig.~\ref{analysis_2}. The clustering analysis reveals notable disparities in the mean PV power generation across these distinct categories. Specifically, commercial customers exhibit the highest PV power generation levels during peak hours, contrasted with residential customers who manifest the lowest PV power generation rates. This is mainly caused by the differences of PV-system cell sizes, which may require separate models for different categories of customers, instead of a single model for the entire dataset.  
	
	\subsection{Result Comparison}
	
	We assess the performance of CM-DNN in prediction tasks over RNN, CNNLSTM, LSTM, and GRU, denoted as CM-RNN, CM-CNNLSTM, CM-LSTM, and CM-GRU, respectively. Comparisons are made with a single predictor trained across RNN, CNNLSTM, LSTM, and GRU to highlight the superiority of CM-DNN.
	
	\begin{figure}[ht!]
		\centering
		\subfigure[]
		{\centering\scalebox{0.5}
			{\includegraphics{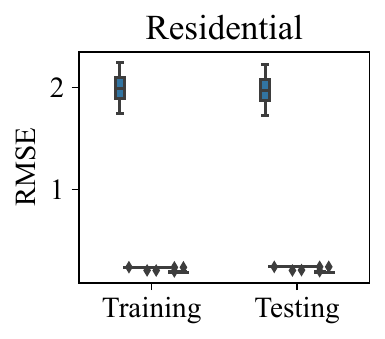}}\label{boxplot_1}} 
		\subfigure[]
		{\centering\scalebox{0.5}
			{\includegraphics{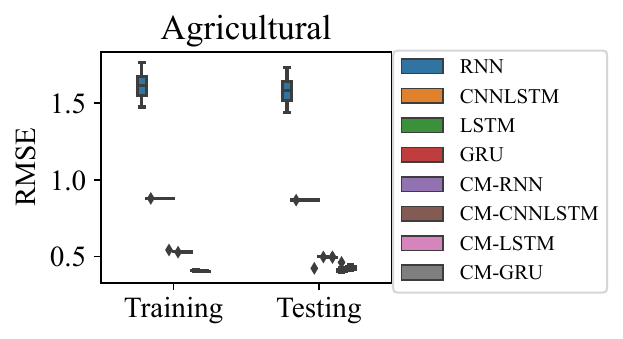}}\label{boxplot_2}} 
		\subfigure[]
		{\centering\scalebox{0.5}
			{\includegraphics{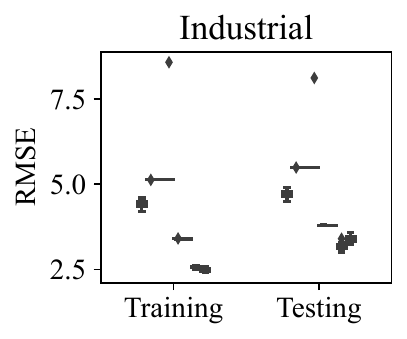}}\label{boxplot_3}} 
		\subfigure[]
		{\centering\scalebox{0.5}
			{\includegraphics{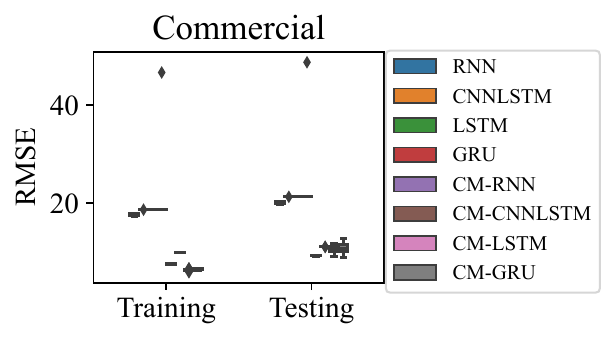}}\label{boxplot_4}} 
		\caption{Boxplots of RNN, CNNLSTM, LSTM, GRU, CM-RNN, CM-CNNLSTM, CM-LSTM, and CM-GRU over training and testing datasets across residential, agricultural, industrial, and commercial datasets.}
		\label{boxplot}
	\end{figure}
	
	\begin{figure}[ht!]
		\centering
		\subfigure[]
		{\centering\scalebox{0.56}
			{\includegraphics{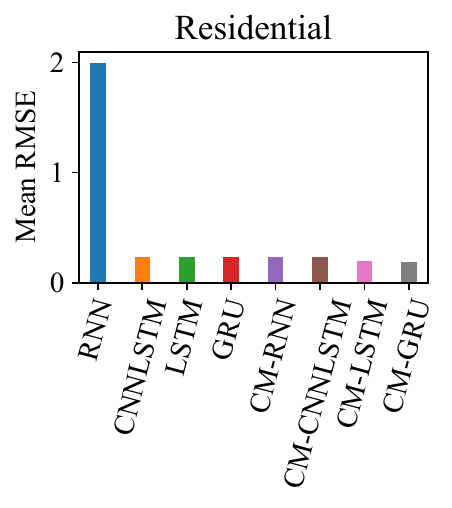}}\label{ave_1}} 
		\subfigure[]
		{\centering\scalebox{0.56}
			{\includegraphics{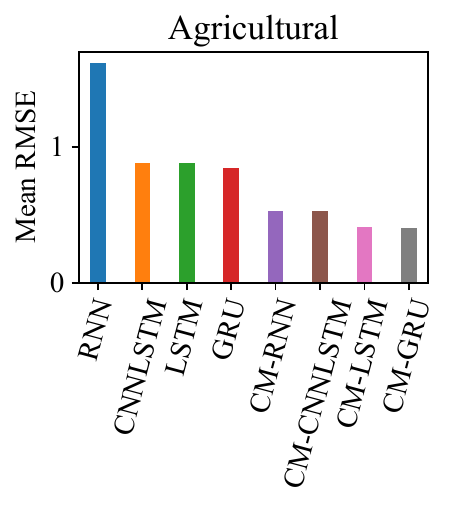}}\label{ave_2}} 
		\subfigure[]
		{\centering\scalebox{0.56}
			{\includegraphics{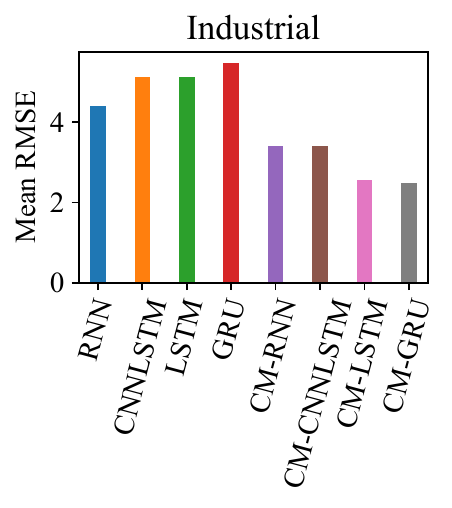}}\label{ave_3}} 
		\subfigure[]
		{\centering\scalebox{0.56}
			{\includegraphics{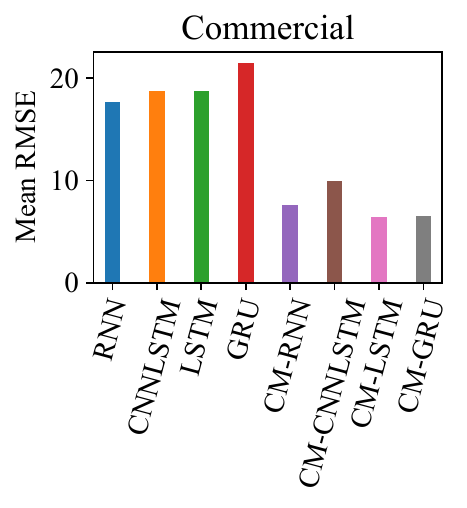}}\label{ave_4}} 
		\caption{Average training RMSE of RNN, CNNLSTM, LSTM, GRU, CM-RNN, CM-CNNLSTM, CM-LSTM, and CM-GRU across residential, agricultural, industrial, and commercial datasets.}
		\label{ave_rmse}
	\end{figure}
	
	\begin{table*}[ht!]
		\centering
		\caption{Average RMSE ± standard deviation for each task over RNN, CNNLSTM, LSTM, GRU, CM-RNN, CM-CNNLSTM, CM-LSTM, and CM-GRU}
		\begin{tabular}{|c|cccc|}
			\hline
			\textbf{Models\textbackslash{}Tasks} & \textbf{Residential}            & \textbf{Agricultural}            & \textbf{Industrial}            & \textbf{Commercial}            \\ \hline
			\textbf{RNN }                                 & 1.9777±0.1573          & 1.5862±0.0916          & 4.7164±0.131           & 20.1362±0.1995         \\
			\textbf{CNNLSTM}                              & 0.2407±0.0             & 0.8689±0.0003          & 5.4919±0.0003          & 21.3023±0.0004         \\
			\textbf{LSTM }                                & 0.2407±0.0001          & 0.8687±0.0007          & 5.4918±0.0006          & 21.3021±0.0007         \\
			\textbf{GRU }                                 & 0.2372±0.0112          & 0.8241±0.1409          & 5.7541±0.8296          & 24.0471±8.6806         \\
			\textbf{CM-RNN}                                & 0.2385±0.0104          & 0.4994±0.0013          & 3.8012±0.0042          & \textbf{9.2927±0.0517} \\
			\textbf{CM-CNNLSTM}                            & 0.2412±0.0001          & 0.4962±0.0001          & 3.7863±0.0012          & 11.1113±0.0185         \\
			\textbf{CM-LSTM}                               & 0.1952±0.0164          & \textbf{0.4156±0.0189} & \textbf{3.1862±0.1144} & 10.4989±0.704          \\
			\textbf{CM-GRU}                               & \textbf{0.1938±0.0167} & 0.4249±0.0135          & 3.3877±0.1052          & 10.7866±1.2032         \\ \hline
		\end{tabular}
		\label{testing}
	\end{table*}
	\begin{figure*}[ht!]
		\centering
		\subfigure[]
		{\centering\scalebox{0.55}
			{\includegraphics{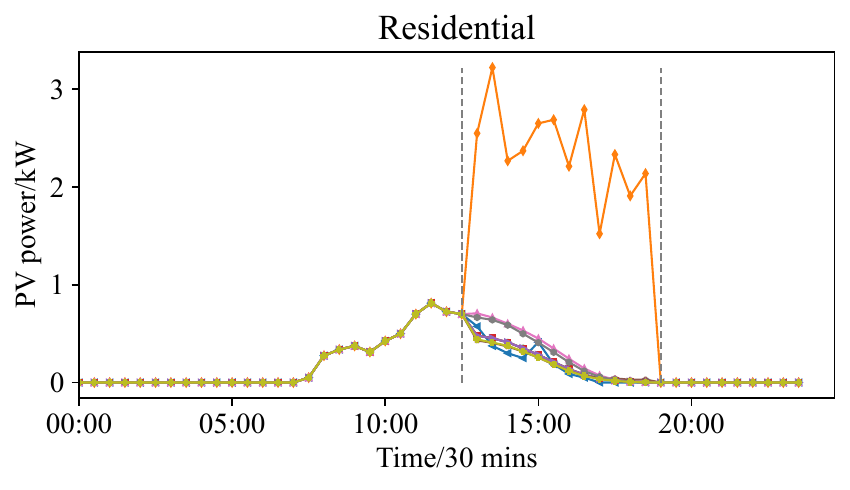}}\label{Vis_1}} 
		\subfigure[]
		{\centering\scalebox{0.55}
			{\includegraphics{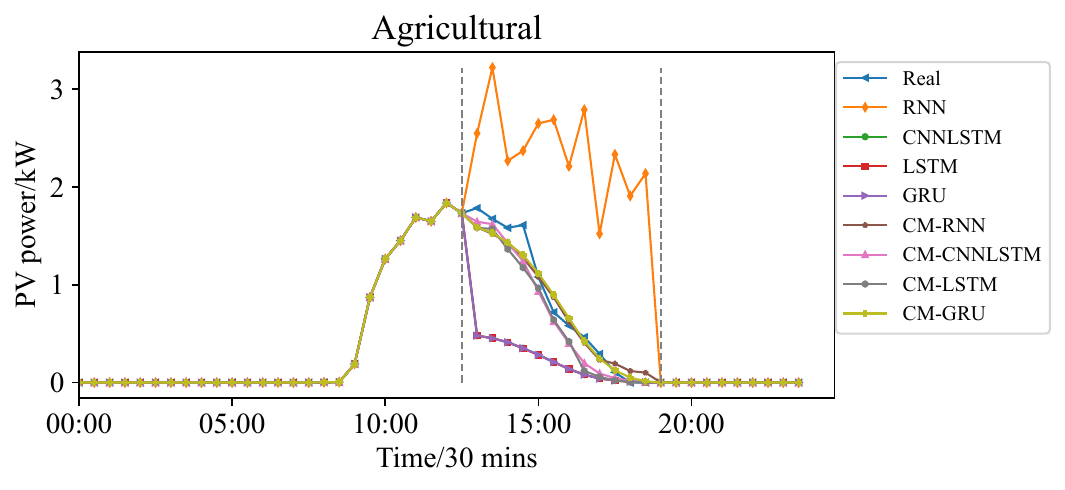}}\label{Vis_2}} 
		\subfigure[]
		{\centering\scalebox{0.55}
			{\includegraphics{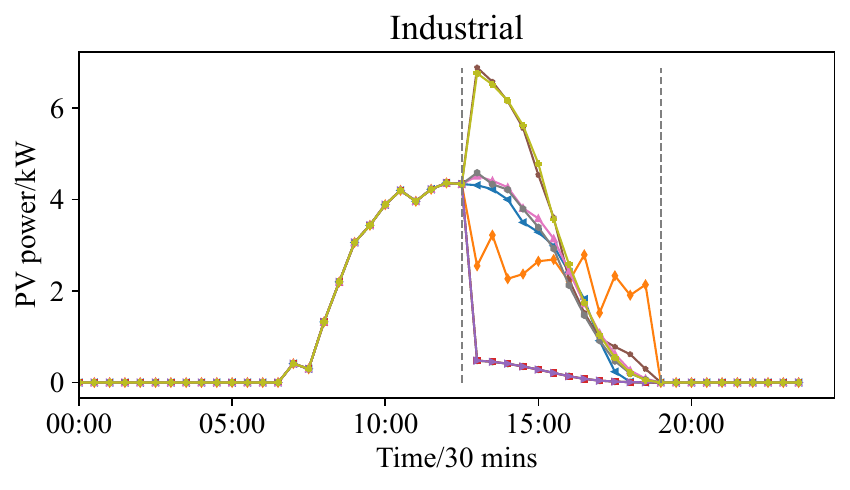}}\label{Vis_3}} 
		\subfigure[]
		{\centering\scalebox{0.55}
			{\includegraphics{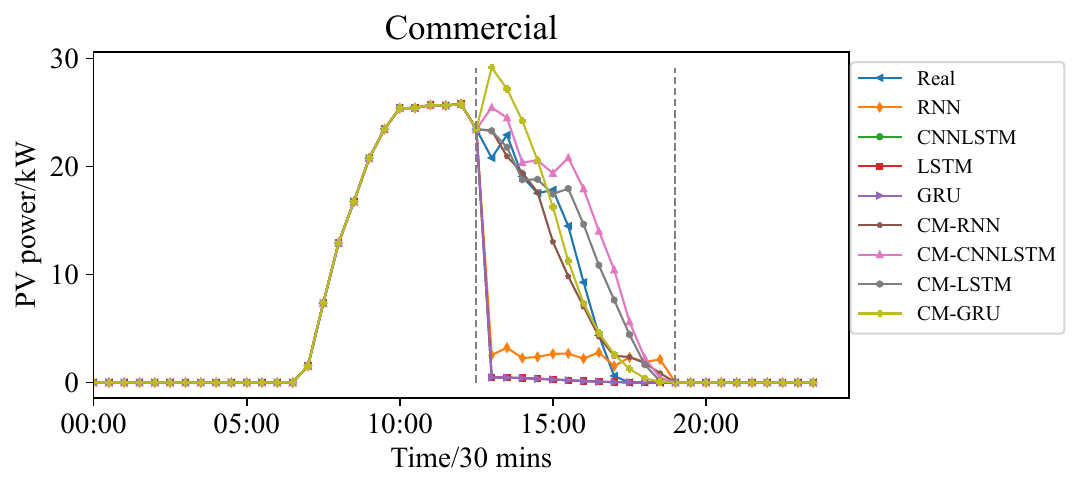}}\label{Vis_4}} 
		\caption{One instance of PV power predictions over testing dataset with RNN, CNNLSTM, LSTM, GRU, CM-RNN, CM-CNNLSTM, CM-LSTM, and CM-GRU across residential, agricultural, industrial, and commercial datasets.}
		\label{Vis}
	\end{figure*}
	
	Fig.~\ref{boxplot} visually depicts the distributions of training and testing RMSE on RNN, CNNLSTM, LSTM, GRU, CM-RNN, CM-CNNLSTM, CM-LSTM, and CM-GRU across residential, agricultural, industrial, and commercial datasets over 10 runs. The outcomes for RNN, CNNLSTM, LSTM, and GRU indicate that each model predicts the entire dataset, encompassing residential, agricultural, industrial, and commercial aspects simultaneously. This allows the model to grasp patterns across all customer categories, transferring and sharing knowledge through connection weights and biases. In~Fig.~\ref{boxplot_1},~Fig.~\ref{boxplot_2},~Fig.~\ref{boxplot_3}, and~Fig.~\ref{boxplot_4}, we observe that our proposed CM-DNN, i.e., CM-RNN, CM-CNNLSTM, CM-LSTM, and CM-GRU, generally performs better than RNN, CNNLSTM, LSTM, and GRU on both training and testing datasets. Specifically,~Fig.~\ref{boxplot_1} and~Fig.~\ref{boxplot_2} reveal that RNN exhibits the poorest prediction performance on residential and agricultural datasets during training. The RMSE distributions in~Fig.~\ref{boxplot_3} and~Fig.~\ref{boxplot_4} illustrate that GRU leads to the lowest accuracy (highest RMSE) than other models on industrial and commercial dataset during training. 
	
	Fig.~\ref{ave_rmse} presents the average training RMSE over 10 independent runs for each model across residential, agricultural, industrial, and commercial datasets. It is evident that the average RMSE over RNN for residential and agricultural datasets in~Fig.~\ref{ave_1} and~Fig.~\ref{ave_2} is the highest, consistent with the findings in~Fig.~\ref{boxplot_1} and~Fig.~\ref{boxplot_2}. Similarly, GRU over industrial and commercial datasets in~Fig.~\ref{ave_3} and~Fig.~\ref{ave_4} verifies the results in~Fig.~\ref{boxplot_3} and~Fig.~\ref{boxplot_4}. Also, for each model, the clustering-based multitasking learning yields higher accuracy than without clustering, e.g., CM-RNN performs better than RNN. Even CM-RNN and CM-CNNLSTM cannot surpass CM-LSTM and CM-GRU, they exhibit lower average RMSE than RNN, CNNLSTM, LSTM, and GRU across all prediction tasks. Furthermore, CM-LSTM and CM-GRU consistently achieve higher accuracy across all training datasets. 
	
	To further demonstrate the superiority of the proposed CM-DNN, for each customer type, we report the mean RMSE associated with standard deviation on the testing dataset, as shown in~Table.~\ref{testing}. The lowest average RMSE (the best prediction performance) is labeled in bold. The results illustrate that CM-DNN generally performs better than RNN, CNNLSTM, LSTM, and GRU on the testing residential, agricultural, industrial, and commercial datasets. For the residential dataset, even CM-CNNLSTM is not comparable to CNNLSTM, CM-GRU outperforms others. CM-LSTM leads to higher accuracy (lower RMSE) than RNN, CNNLSTM, LSTM, GRU, CM-RNN, CM-CNNLSTM, and CM-GRU for agricultural and industrial datasets. CM-RNN exhibits the highest prediction performance than other models on commercial dataset. Overall, except CM-CNNLSTM on residential dataset, CM-DNN across RNN, CNNLSTM, LSTM, and GRU outperforms the predictor without clustering for all tasks, further verifying the findings in~Figs.~\ref{boxplot} and~\ref{ave_rmse}.
	
	\subsection{Visualizations}
	
	To visualize how the predicted values of each model is similar to the real values, we show one of the instances on testing dataset across residential, agricultural, industrial, and commercial in~Fig.~\ref{Vis}. For each instance, the data between 6:00 am and 1:00 pm serves as the input, and the related output is from 1:30 pm to 6:30 pm (between the gray vertical lines). For the residential dataset in~Fig.~\ref{Vis_1}, RNN exhibits worse testing performance since its prediction values deviate significantly from the real values.~Fig.~\ref{Vis_2} shows that the predicted values over RNN, LSTM, and GRU are significantly different from the real values. The predictions over industrial dataset in~Fig.~\ref{Vis_3} indicate that RNN, LSTM, GRU, CM-RNN, CM-GRU lead to worse results. RNN, LSTM, GRU, and CM-GRU do not predict accurately on commercial dataset in~Fig.~\ref{Vis_4}. These findings in~Fig.~\ref{Vis} further verify the results in~Fig.~\ref{boxplot} and Table.~\ref{testing}. 
	
	\section{Conclusions and Future Work}\label{conclusion}

In this paper, we proposed a clustering-based multitasking deep neural network (CM-DNN) framework to address PV power generation forecasting. The whole dataset is de-identified with several categories of customers. To ensure accurate predictions, K-means was applied to cluster the entire dataset into residential, agricultural, industrial, and commercial datasets based on data distributions. DNNs were first trained on each dataset using a GD-based optimizer. For these trained DNNs, inter-model knowledge transfer was designed to further enhance their accuracies. Specifically, source task selection was executed to determine the optimal subset of source tasks. Each selected task used a coefficient to decide the amount of DNN model information to transfer and reuse for the target task that corresponds to one of the clusters. To obtain the optimal subset and coefficients, PSO was employed. CM-DNN was tested on several popular time series prediction models such as RNN, CNNLSTM, LSTM, and GRU. To demonstrate the superiority of CM-DNN, the prediction accuracy from the entire PV power generation dataset, trained with a single predictor, was compared. The results showed that inter-model knowledge transfer could lead to higher prediction accuracy than intra-model self-learning, i.e., without clustering. In the future, we will investigate the influence of temperature on the model, study supervised clustering~\cite{qin2005initialization} based on meta-information and other EAs~\cite{qin2013differential} in the proposed CM-DNN framework, and apply CM-DNN to other prediction tasks~\cite{gao2022location}.
	
	\bibliographystyle{IEEEtran}
	\bibliography{Refs}
	
\end{document}